\documentclass{article}
\usepackage{spconf,amsmath,graphicx}
\usepackage{amssymb}
\usepackage{url}

\newcommand{\ie}{\textit{i.e.}}
\newcommand{\eg}{\textit{e.g.}}

\title{PersonificationNet: Making customized subject act like a person}
\name{Tianchu Guo, Pengyu Li, Biao Wang, Xiansheng Hua* \thanks{*Corresponding author, Fellow of IEEE, Email: xshua@outlook.com.}}
\address{Terminus Labs, Terminus Group, China}
\begin{document}
%
\maketitle
\begin{abstract}
Recently customized generation has significant potential, which uses as few as 3-5 user-provided images to train a model to synthesize new images of a specified subject. Though subsequent applications enhance the flexibility and diversity of customized generation, fine-grained control over the given subject acting like the person's pose is still lack of study. In this paper, we propose a PersonificationNet, which can control the specified subject such as a cartoon character or plush toy to act the same pose as a given referenced person's image. It contains a customized branch, a pose condition branch and a structure alignment module. Specifically, first, the customized branch mimics specified subject appearance. Second, the pose condition branch transfers the body structure information from the human to variant instances. Last, the structure alignment module bridges the structure gap between human and specified subject in the inference stage.
Experimental results show our proposed PersonificationNet outperforms the state-of-the-art methods.
\end{abstract}
\section{Introduction}
\label{sec:intro}
Large text-to-image models~\cite{nichol2021glide,dhariwal2021diffusion,kim2022diffusionclip,brooks2023instructpix2pix,kawar2023imagic,hertz2022prompt} have demonstrated remarkable capabilities in generating high-quality and diverse images. Based on this technique,
making specified subject displayed in a new scene has been realized by the customized generation, such as Dreambooth~\cite{ruiz2023dreambooth}, custom diffusion~\cite{kumari2023multi}, and Textual Inversion~\cite{gal2022textualInversion}, which only uses 3-5 user-provided images to inject the subject into the model and synthesize novel renditions within various contexts. They make the user synthesized images of their own concepts with cost-effective training processes.

\begin{figure}[!t]
	\centering
	\includegraphics[width=0.5\textwidth]{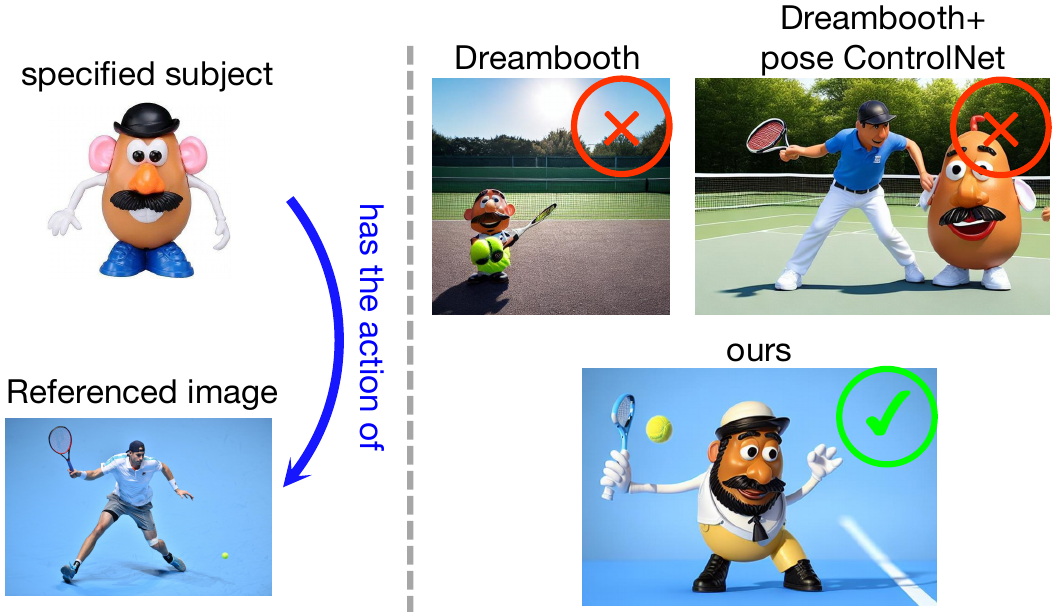}
	\caption{ 
	Problem definition and challenge. The left part shows given the specified Mr. Potato images, we try to let Mr. Potato replicate the actions of the referenced image. The right part shows that existing method struggle to achieve this goal but our proposed method succeeds to generate.}
	\label{fig:intro}
 \vspace{-15pt}
\end{figure}

However, existing methods can not make the specified subject act as the referenced images, \eg both the same pose and the same background, as shown in Fig. \ref{fig:intro}. Given the specified subject, \eg, the ``Mr. Potato'',  Dreambooth~\cite{ruiz2023dreambooth} mimics its appearance, fails to synthesize the same pose with the referenced images. It means only using the text prompt doesn't satisfy our needs. Dreambooth combined with ControlNet~\cite{zhang2023controlnet} which utilizes the human pose image as additional condition still gives a bad generation. Though it has elements of the ``Mr. Potato'' and ``playing tennis'', the contents is not the ``Mr. Potato is playing tennis''. The main problems are, (1) the pose condition branch of the original ControlNet inherently relies on prior knowledge of human body structure, which cannot effectively control the pose of "Mr. Potato", (2) the pose condition image extracted from the referenced image has large structure gap between the human body and the specified subject, and this mismatch further adds complexity to the generation process.

To solve these problems, in this paper, the PersonificationNet is proposed, which contains a customized branch, a pose condition branch and a structure alignment module. Specifically, the customized branch mimics the specified subject appearance by trained on 3-5 user provided images. The pose condition branch transfers the body structure information from the human to variant instances by finetuning with our collected dataset. The structure alignment module makes the pose image has the body structure of the given subject and the pose of the referenced image, bridging the structure gap between human and specified subject in the inference stage. Experimental results show our proposed PersonificationNet outperforms the state-of-the-art methods.

\section{Related Work}
\label{sec:format}
Generative models~\cite{rombach2022SD,saharia2022imagen,ramesh2022dalle} have recently bestowed us with remarkable capabilities, enabling the creation of high-quality images. As the text prompt input can not describe the details such as the subjects in user-given set, customized generation~\cite{cohen2022PALAVRA,gal2022textualInversion,ruiz2023dreambooth} is proposed, which injects new concepts into the original Stable Diffusion Model~\cite{rombach2022SD}. 
Specifically, Textual Inversion~\cite{gal2022textualInversion} introduces new text tokens to learn new concepts, and Dreambooth~\cite{ruiz2023dreambooth} employs rare token to mimic the subject's appearance. Both of them support inpainting and provide an important way to control the final results.

ControlNet~\cite{zhang2023controlnet} utilizes another branch as additional input to control the original stable diffusion model to generate images. Specifically, the control branch has different type such as Canny Edge, human pose, depth and so on.

\section{Method}
\label{sec:pagestyle}
\begin{figure}[!t]
	\centering
	\includegraphics[width=0.5\textwidth]{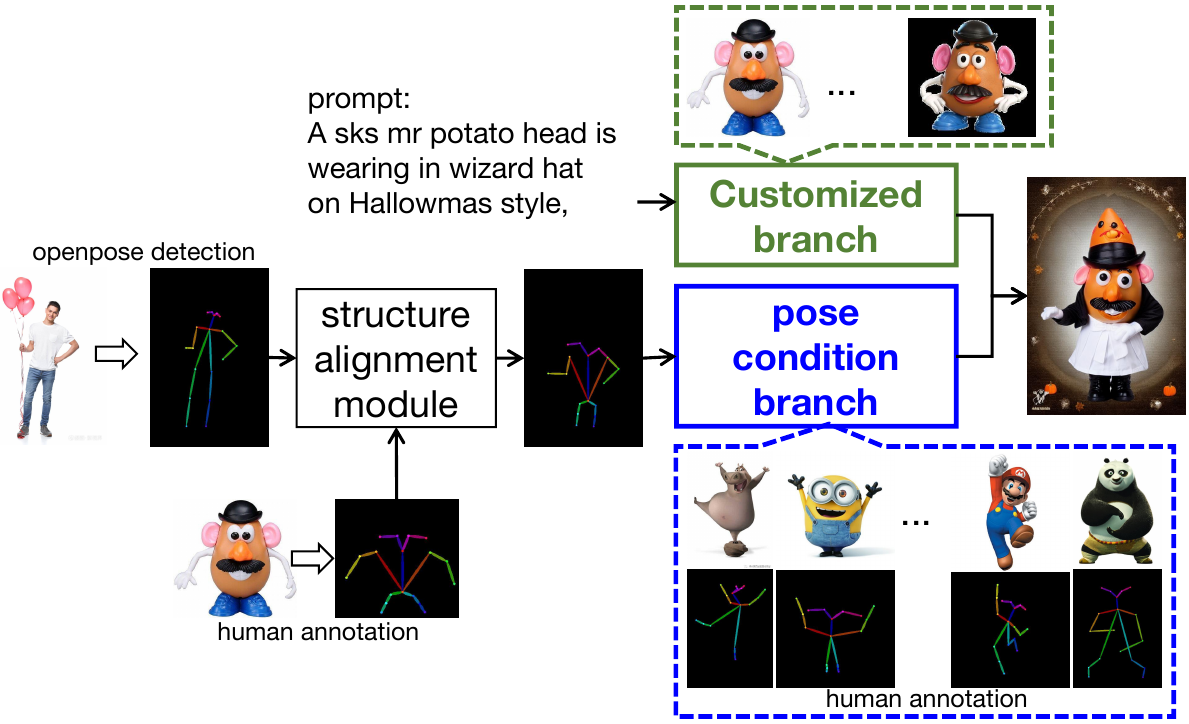}
	\caption{ 
	Pipeline of the PersonificationNet. }
	\label{fig:method}
\end{figure}

The proposed PersonificationNet contains a customized branch, a pose condition branch, and a structure alignment module, as shown in Fig.\ref{fig:method}. Specifically, the customized branch and the pose condition branch are trained independently. The structure alignment module is used in the inference phase. Details will be described in the next subsections.


\subsection{Customized Branch}
The customized branch is trained with the 3-5 user-provided images in $<$image, text$>$ pairs.

\textbf{Image-text pairs.}
Given 3-5 images of a specified subject, the BLIP~\cite{li2023blip} model is used to generate the accompanying text captions. Subsequently, we replace the subject of the sentence with a unique identifier, which will be elaborated upon in the following section.

\textbf{Unique identifier selection.}
The unique identifier consist of a rare-token and some common tokens.
The rare-token similar to Dreambooth~\cite{ruiz2023dreambooth} is used to represent the specified concepts. The common tokens describe most characteristic of the specified subject. For example, we use ``sks Mr. potato head'' as the unique identifier, where the ``sks'' serves as  the rare-token and ``Mr. potato head'' represents the multiple common tokens. 
 
\textbf{Training scheduler.}
The training scheduler is similar to Kumari's method~\cite{kumari2023multi}.
Given the $<$image, text$>$ pairs, the customized branch, which is a stable diffusion model, will be optimized through the approximation of the original data distribution $q(x_0)$ with $p(x_0)$:
\begin{equation}
    p(\textbf{x}_0)=\int [p_{\theta}(\textbf{x}_T)\prod p_{\theta}^t(\textbf{x}_{t-1}\mid \textbf{x}_{t})]d\textbf{x}_{1:T} \qquad
\end{equation}
where $x_1$ to $x_T$ extracted by VAE~\cite{rombach2022SD} encoder are latent variables of a forward Markov chain, and s.t. $x_t$ = $\sqrt{\alpha_{t}}x_0+\sqrt{1-\alpha_t} \epsilon$. The length of the Markov chain often fixed to 1000. When presented with a noisy image, denoted as $x_t$ at timestep $t$, the model learns to denoise the input image to obtain $x_{t-1}$. The training objective of the model follows,
\begin{equation}
     \mathcal{L} = \mathbb{E}_{\epsilon,\textbf{x},\textbf{c},t}[\omega_{t} \parallel \epsilon-\epsilon_{\theta}(\textbf{x}_t,\textbf{c},t) \parallel^2_2]
\end{equation}
where $\epsilon_{\theta}$ is the model prediction and $\omega_{t}$ is a time-dependent weight on loss. The model is conditioned on the timestep $t$ and text prompt $c$. 
During inference, a random Gaussian latent $x_T$ is denoised for fixed timesteps using the model. Subsequently, the denoised latent variable is decoded by a Variational Autoencoder (VAE) to generate an image.

\begin{figure*}[!t]
	\centering
    	\includegraphics[width=0.95\textwidth]{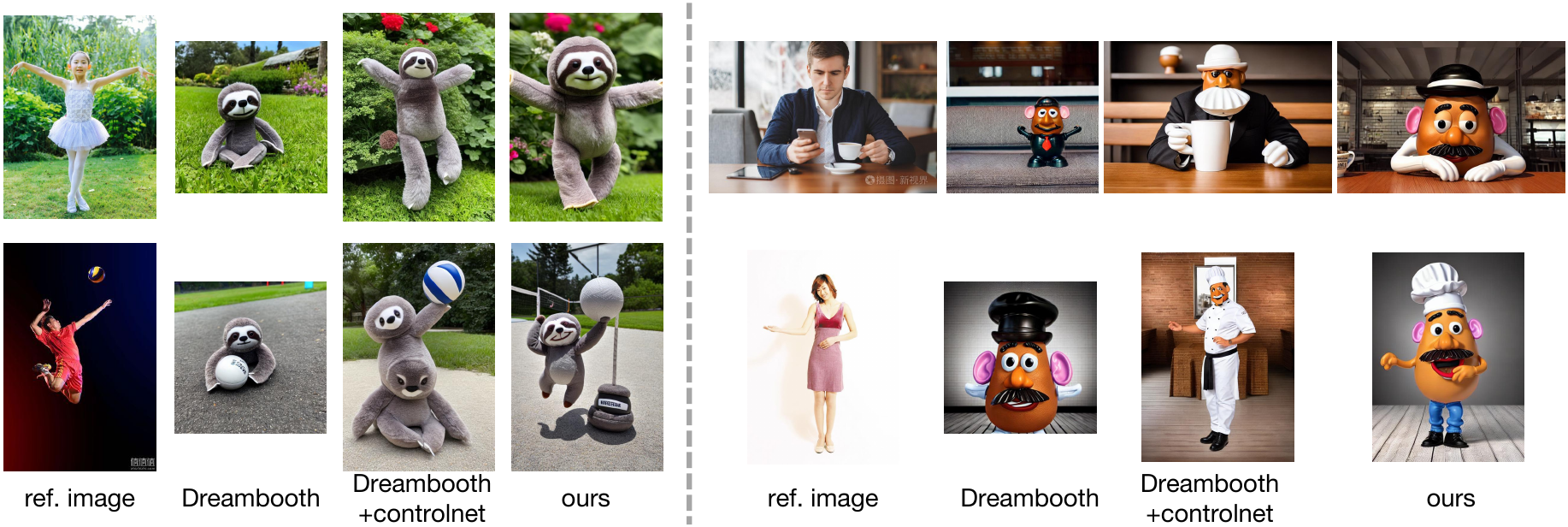}
	\caption{ 
	Comparison with existing methods.}
	\label{fig:sota}
  \vspace{-15pt}
\end{figure*}

\subsection{Pose Condition Branch}
The pose condition branch has the same architecture described in~\cite{zhang2023controlnet}. It provides pose condition on the neural network blocks to control the customized branch following
\begin{equation}
    y_c = \mathcal{F}(x;\Theta)+\mathcal{Z}(\mathcal{F}(x+\mathcal{Z}(c;\Theta_{z1});\Theta_c);\Theta_{z2})
\end{equation}
where the ${\Theta_{z1},\Theta_{z2},\Theta_{c}}$ are the parameters of the pose condition branch and $\Theta$ is the parameters of the customized branch. The $\mathcal{F}$ is the neural network block, the $\mathcal{Z}$ is the pose condition branch convolution based operation and $y_c$ is the block's output with pose condition.

The original parameters of pose condition branch in~\cite{zhang2023controlnet} are trained with human pose images, which primarily captured the prior of human body structure scale, leading to mismatch when applied to the body structure of the specified subject. Thus, the pose condition branch needs to be finetuned with our collected images, which has 55 images with different body structure, aiming to transfer the body structure information from the human's to that of variant subjects.

The dataset is collected on website whose subjects has variant body structure. The skeleton points of the collected images are annotated manually. Then the pose images are drawn according to the skeleton points, to making the triplet pair $<$image, text, pose image$>$. The weights of the pose condition branch is initialed with the weights of pose control net, which has the human body structure prior. These initial weights were then fine-tuned to adapt to the new body structure characteristic of the specified subject, reducing the body structure gap between the human and the specified subject.

\subsection{Structure Alignment Module}
The structure alignment module further reduces the structure gap between the referenced human image and the specified subject in the inference stage. It contains four steps.

First, two virtual points, \ie the shoulder center and the hip center, are computed as shoulder and hip points' father nodes, who will be used in the next part.

Second, the skeleton points Euclidean coordinates, \ie $(x,y)$, will be converted to polar coordinates,\ie $(\rho, \cos\theta, \sin\theta)$. These polar coordinates are calculated relative to their respective father. Specifically, the elbows' father nodes are the shoulders, the wrists' father nodes are the elbows. Note that the shoulders' father nodes are the virtual point shoulder center, and the hips' father nodes are the hip center.

Third, aligned points follows
\begin{equation}
    pos_{ali.}^i=(\rho_{sub.}^i,\cos\theta_{ref.}^i,\sin\theta_{ref.}^i)
\end{equation}
where the $ali.$,$ref.$ and $sub.$ mean the coordinates of aligned, referenced and specified subject, respectively. The $i$ means the index of the points.
The aligned points not only preserve the body proportions of the specified subject but also have the pose of the referenced image, achieving a harmonious blend of these two key aspects.

Last, the aligned points will be convert to Euclidean space. To ensure that the transformed skeleton matches the position of the referenced skeleton, a correction is applied, computed as: $\hat{y}_{ali.} = y_{ali.} + b$, where the $b$ is equal to $min(y_{ref.}^i) - min(y_{ali.}^i),i\in {0,1...N}$. The $N$ is the points number defined by openpose~\cite{cao2017openpose}.

The pose images drawn according to the $( x,\hat{y} )_{ali.}$ is the input of the pose condition branch.

\section{Experiments}
\label{sec:Experiments}
\subsection{Settings}
The customized branch was trained with the learning rate 1e-5 for 400 iterations. Then the customized branch was fixed to train the pose condition branch using our collected dataset with the learning rate 1e-5 for 800 iterations.

Two specified subjects are employed to evaluate, \ie the ``Mr. Potato'' and the ``sloth''.
The ``Mr. Potato'' comes from the internet and its model weights was downloaded\url{https://huggingface.co/sd-Dreambooth-library/mr-potato-head} for the customized branch with the corresponding unique identifier ``sks mr. potato head''.
The ``sloth'' comes from the open source dataset in~\cite{ruiz2023dreambooth}. 
The unique identifier is ``sks grey sloth plushie''. 

The initial weights of the pose condition branch was downloaded from HuggingFace~\cite{HuggingFace}.


Existing methods, \ie Dreambooth~\cite{ruiz2023dreambooth} and Dreambooth combined with pose condition ControlNet~\cite{zhang2023controlnet} were employed to evaluate. Specifically, Dreambooth~\cite{ruiz2023dreambooth} customized the specified subject and utilizes text prompt to generate images. Dreambooth combined with pose condition ControlNet~\cite{zhang2023controlnet} additionally add pose image extracted from the referenced images.

\begin{figure*}[!tp]
	\centering
	\includegraphics[width=0.95\textwidth]{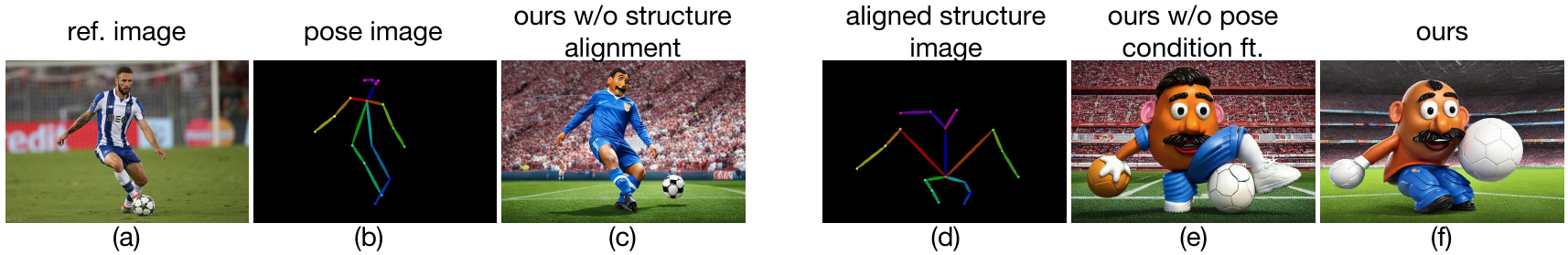}
	\caption{ 
	Results of ablation study.}
	\label{fig:ablation}

\end{figure*}

\begin{figure*}[!tp]
	\centering
	\includegraphics[width=0.95\textwidth]{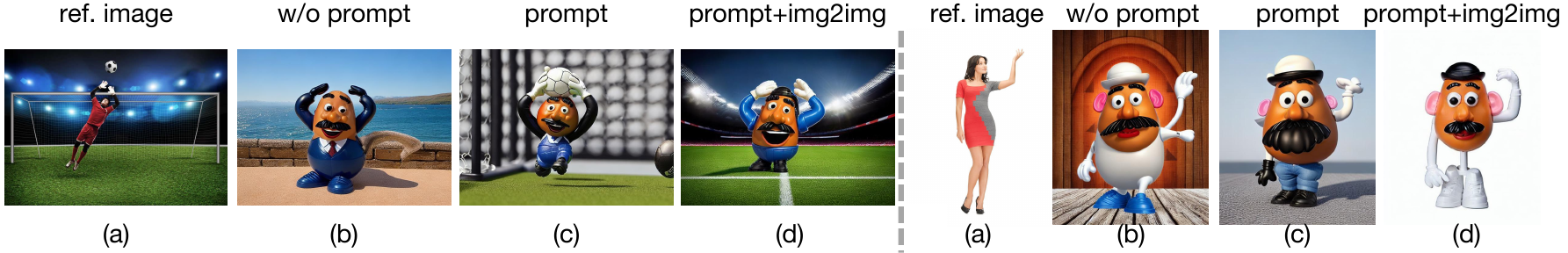}
	\caption{ 
	Variant applications of the PersonificationNet.}
	\label{fig:variation}
  \vspace{-15pt}
\end{figure*}

\vspace{-7pt}
\subsection{Comparison with Existing Methods}
The results compared with existing methods are shown in Fig.\ref{fig:sota}. The left part shows the results of the ``sloth'' and the right part shows the results of the ``Mr. Potato''. 

It can be seen that, Dreambooth which only uses text prompt can not satisfy the needs, though this method mimics the appearance of the specified subjects. 

Dreambooth+ControlNet method utilizes pose image extracted from the referenced image as the additional condition input, but still got a bad result. It is because the pose condition ControlNet learns the human's structure scale, which is not fit to the specified subject. 
As shown in the right part of the Fig.\ref{fig:sota}, though the generated images exhibit the same pose with the referenced image,  ``Mr. Potato'' has the human's body and ``Mr. Potato'' 's face. The action and the appearance can't be combined harmoniously. 
 
The proposed method, denoted as ``ours'' in Fig.\ref{fig:sota}
 shows that the images generated by our method not only keep the appearance of the specified subject, but also act the same pose as the referenced images.

\vspace{-7pt}
\subsection{Ablation Study}
In the proposed method, both the fintuned pose condition branch and the structure alignment module aim to reduce the body structure gap between the human and the specified subject. This part will discuss the effect of these parts. 

As shown in Fig.\ref{fig:ablation}, (a) represents the referenced image, and its corresponding pose figure is depicted in (b). The (c) is the result generated by the proposed method without the structure alignment module.  It's evident that the generated image still has the human's structure, which is disharmonious with the specified ``Mr. Potato''. 

(d) represents the pose image processed by the structure alignment module, which obtains a harmonious combination of the body structure of the specified subject and the action pose of the referenced image. The (e) is the image generated without the pose condition branch finetuning. Compared with (c), the (e) exhibits "Mr. Potato"'s body structure, but artifacts are still noticeable on its legs. It is because the pose condition branch without finetuning struggles to adapt to the specified pose structure, resulting in inconsistencies. The (f) is generated by the method with both pose condition branch finetuning and structure alignment module. It can be seen that the pose structure gap has significantly reduced by our proposed method. The result is a more coherent and harmonious image that accurately represents the specified subject appearance while keeping the pose of the referenced image.

\vspace{-7pt}
\subsection{Variant Applications of the PersonificationNet}
In this part, the variant application of the proposed method will be tested. 
As shown in Fig.\ref{fig:variation}, the left part and the right part shows cases from two referenced images. The text prompts are ``The sks mr. potato head is a soccer goalkeeper catching the ball in the air'' and ``The sks mr. potato head is standing'', respectively.

Specifically, the generated images shown in (b) exhibit the same pose as the referenced images, even in the absence of a text prompt. 
With the input of a text prompt, the generated images shown in (c) not only maintain the same pose but also accurately depict the specified action. 

Furthermore, The result in (d) employs a higher level of condition information, incorporating both the text prompt and the referenced image as conditions. It means in the inference process, the latent $x_T$ is the noisy embedding of the referenced image encoded by VAE's encoder, replacing the random Gaussian noise. It can be seen that the generated image in (d) not only has the same pose and action, but also has the same background with the referenced image.

From (b) to (d), the condition information to the model is continuously increased, and the quality of the images generated by the model and the similarity with the referenced images are also getting better. 
\vspace{-7pt}
\section{Conclusion}
\label{sec:Conclusion}
In this paper, we point out that existing methods fail to make the user provided subject act as the same pose and background with the referenced images. It is because there is great structure gap between the human body and the specified subject. To solve this problem, the PersonificationNet is proposed, which contains a customized branch, a pose condition branch, and a structure alignment module. Experimental results show that the proposed method outperforms the existing methods.
\bibliographystyle{IEEEbib}
\bibliography{refs}

\end{document}